\documentclass[letterpaper, 10 pt, journal, twoside]{IEEEtran}

\usepackage[T1]{fontenc}
\usepackage{cite}
\usepackage{amssymb,amsfonts}
\usepackage{blindtext}
\makeatletter
\let\NAT@parse\undefined
\makeatother
\usepackage{hyperref}
\usepackage{graphicx}
\usepackage{textcomp}
\usepackage{mathtools}
\usepackage{changes}
\usepackage{subcaption}
\usepackage{algorithm}
\usepackage{xcolor}
\usepackage{color, soul}
\usepackage{amsmath}
\usepackage{booktabs}
\usepackage{multirow}
\usepackage{xstring}
\usepackage{hhline}
\usepackage{kotex}
\usepackage{siunitx}
\usepackage{comment}
\usepackage{physics}
\usepackage{tikz}
\usepackage{cleveref}
\usepackage{lipsum}
\usepackage{bm}
\usepackage{siunitx}
\usepackage{array}
\usepackage{eqparbox}
\usepackage{caption}
\usepackage{amsthm}
\usepackage[noend]{algorithmic}

\usepackage{etoolbox}  
\makeatletter
\patchcmd{\algorithmic}{\addtolength{\ALC@tlm}{\leftmargin} }{\addtolength{\ALC@tlm}{\leftmargin}}{}{}
\makeatother

\crefformat{section}{\S#2#1#3} 
\crefformat{subsection}{\S#2#1#3}
\crefformat{subsubsection}{\S#2#1#3}

\floatname{algorithm}{Algorithm}

\definecolor{rv}{RGB}{0, 0, 255}
\theoremstyle{definition}

\sethlcolor{lightgray}

\makeatletter
\DeclareRobustCommand{\iscircle}{\mathord{\mathpalette\is@circle\relax}}
\newcommand\is@circle[2]{%
  \begingroup
  \sbox\z@{\raisebox{\depth}{$\m@th#1\bigcirc$}}%
  \sbox\tw@{$#1\square$}%
  \resizebox{!}{\ht\tw@}{\usebox{\z@}}%
  \endgroup
}
\makeatother

\IEEEoverridecommandlockouts                              

\usepackage{amsmath} 
\usepackage{caption}
\newcommand{\rom}[1]{\uppercase\expandafter{\romannumeral #1\relax}}
\title{\LARGE \bf
Retro-RL: Reinforcing Nominal Controller With Deep Reinforcement Learning for Tilting-Rotor Drones}

\author{I Made Aswin Nahrendra$^{1}$, Christian Tirtawardhana$^{1}$, Byeongho Yu$^{1}$, \\Eungchang Mason Lee$^{1}$ and Hyun Myung$^{1,*}$, \textit{Senior Member, IEEE}
\thanks{Manuscript received: February 24, 2022; Revised: May 20, 2022; Accepted: June 16, 2022. 
This paper was recommended for publication by Editor Jens Kober upon evaluation of the Associate Editor and Reviewers' comments. This work was supported by the National Research Foundation of Korea (NRF) Grant funded by the Ministry of Science and ICT for First-Mover Program for Accelerating Disruptive Technology Development (NRF-2018M3C1B9088328). The students are supported by BK21 FOUR.}

\thanks{$^1$The authors are with the School of Electrical Engineering at Korea Advanced Institute of Science and Technology (KAIST), Daejeon, 34141, Republic of Korea. {\tt\small \{anahrendra, christiant, bhyu, eungchang\_mason, hmyung\}@kaist.ac.kr}. \hfill \break
\indent $^*$Corresponding author: Hyun Myung \hfill \break
\indent Digital Object Identifier (DOI): see top of this page.
}
}

\begin{document}

\markboth{IEEE Robotics and Automation Letters. Preprint Version. June 2022}
{Nahrendra \MakeLowercase{\textit{et al.}}: Retro-RL: Reinforcing Nominal Controller with Deep Reinforcement Learning for Tilting-Rotor Drones} 
\maketitle
\IEEEpeerreviewmaketitle

\newtheorem{theorem}{Theorem}
\begin{abstract}
Studies that broaden drone applications into complex tasks require a stable control framework. Recently, deep reinforcement learning (RL) algorithms have been exploited in many studies for robot control to accomplish complex tasks. Unfortunately, deep RL algorithms might not be suitable for being deployed directly into a real-world robot platform due to the difficulty in interpreting the learned policy and lack of stability guarantee, especially for a complex task such as a wall-climbing drone. This paper proposes a novel hybrid architecture that reinforces a nominal controller with a robust policy learned using a model-free deep RL algorithm. The proposed architecture employs an uncertainty-aware control mixer to preserve guaranteed stability of a nominal controller while using the extended robust performance of the learned policy. The policy is trained in a simulated environment with thousands of domain randomizations to achieve robust performance over diverse uncertainties. The performance of the proposed method was verified through real-world experiments and then compared with a conventional controller and the state-of-the-art learning-based controller trained with a vanilla deep RL algorithm.
\end{abstract}


\section{Introduction} \label{sec:intro}

\begin{figure}[t!]
    \begin{subfigure}[]{0.46\textwidth}
		{\label{figure:retro-rl_framework}\includegraphics[width=1.0\textwidth]{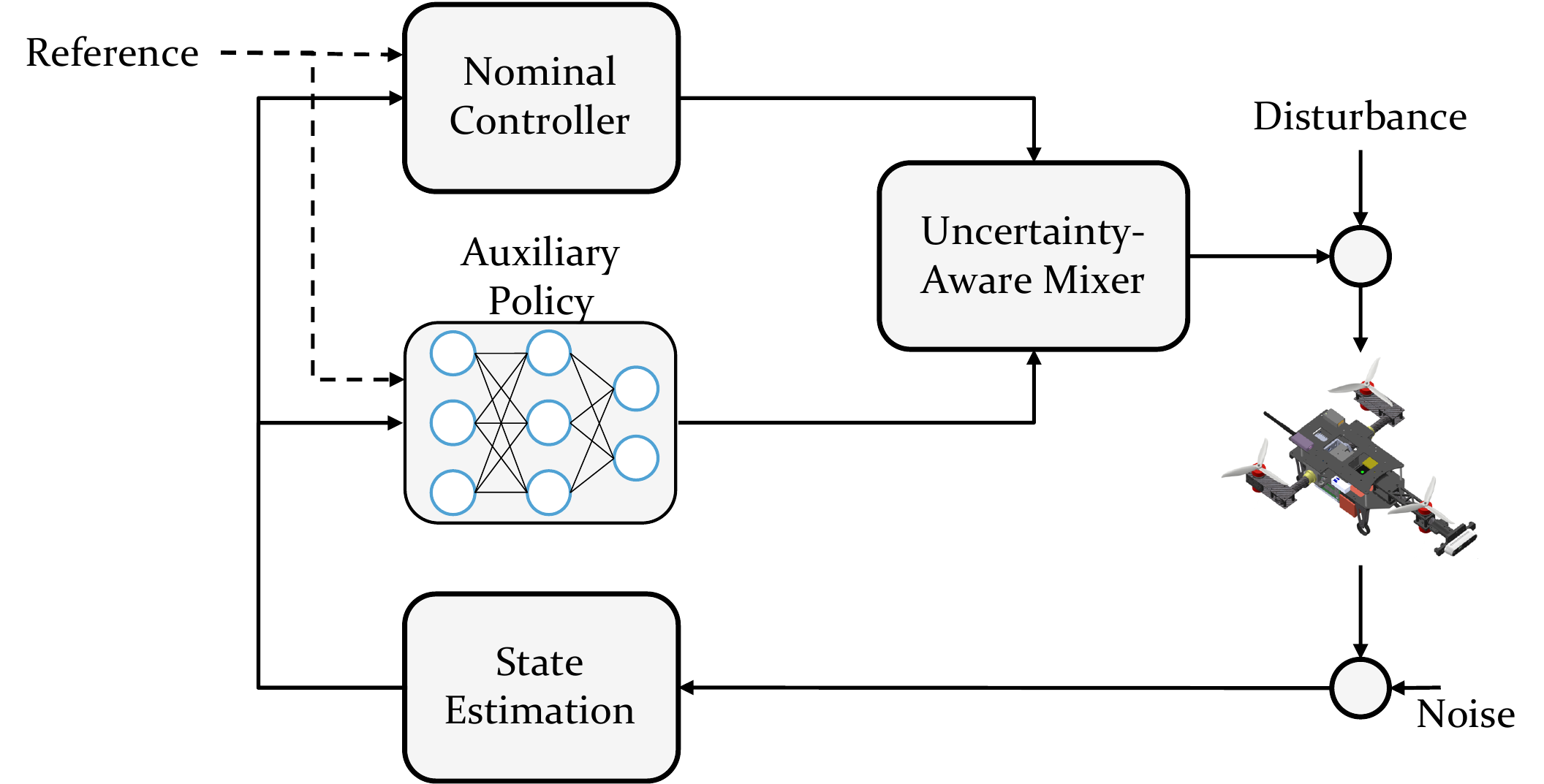}}
		\caption{}
	\end{subfigure}
	\begin{subfigure}[]{0.46\textwidth}
		\includegraphics[width=1.0\textwidth]{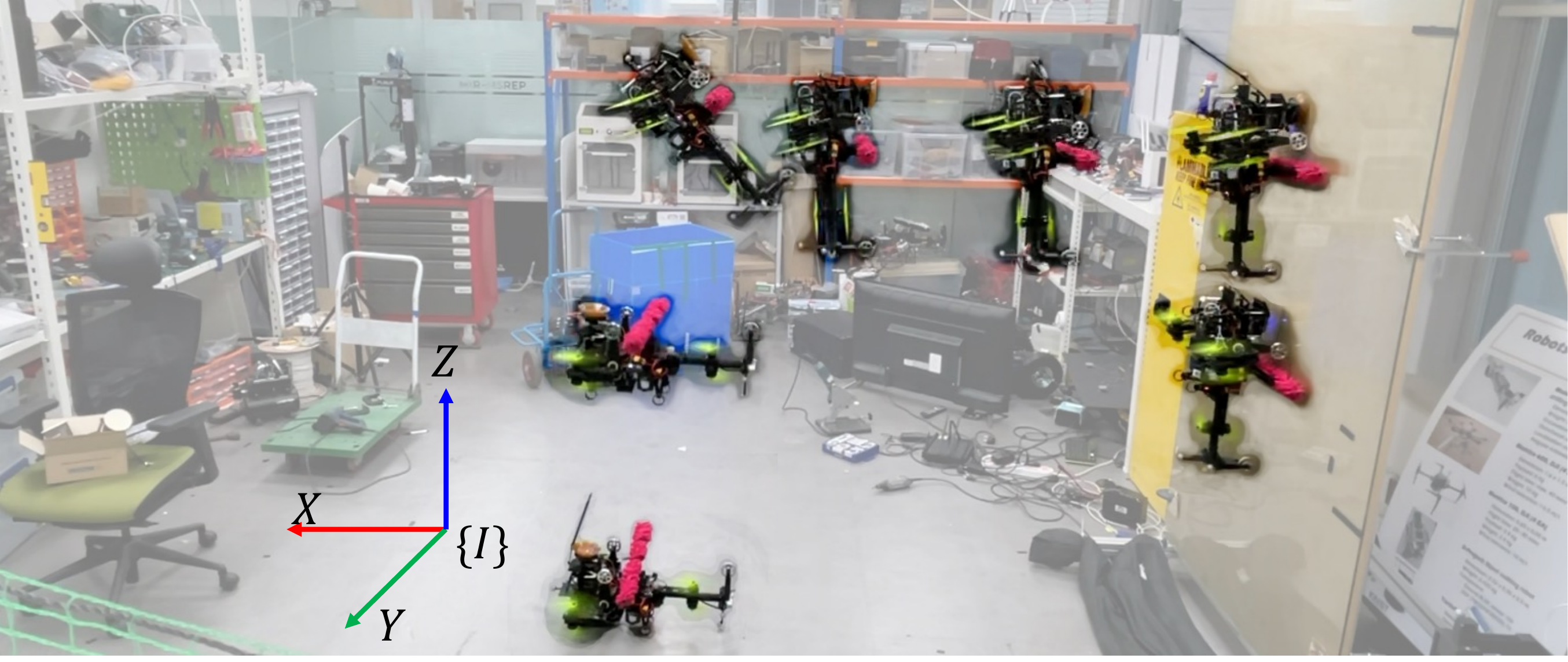}
		\caption{}\label{figure:wall_climbing}
	\end{subfigure}
	\captionsetup{font=footnotesize}
    \caption{(a) The control framework consists of a hybrid structure for mixing a stable nominal controller with a robust performance RL policy with an uncertainty-aware control mixer. (b) The superiority of Retro-RL is validated on CAROS-Q~\cite{lee2021caros}.} 
	\label{fig:overview}
	\vspace{-0.3cm}
\end{figure}

\IEEEPARstart{T}{ilting-rotor} drones have gained significant interest in recent years due to their potential applications in urban areas~\cite{youn2021collision, kim2017robust, myeong2018development}. The applications include structural inspection~\cite{myeong2018development, myeong2019development}, contact-based wall-cleaning~\cite{lee2021caros}, and 6-DoF manipulation~\cite{kamel2018voliro}. These atypical configurations offer advantages in terms of omnidirectionality~\cite{kamel2018voliro} and prevention of collision between propellers and the wall for close-wall inspection~\cite{lee2021caros}. However, those atypical configurations of tilting-rotor drones pose significant challenges in their controller design. The fundamental solution for controlling such multirotor platforms is via meticulous system modeling that results in calculating a dynamic allocation matrix to control individual rotors~\cite{kamel2018voliro,lee2021caros,falanga2018foldable,kim2021morphing,allenspach2020design}, referred to as a nominal controller in this paper. Nominal controllers, however, are prone to difficulties due to real-world uncertainties, for instance, disturbances due to the counteracting airflow, known as wall effect~\cite{davis2018aerodynamic,kocer2018centralized}, and interaction with a structure~\cite{lee2020aerial}. Therefore,  a control strategy with robust performance is required to enhance the performance in more complex scenarios such as wall-climbing (Fig.~\ref{fig:overview}).

In contrast with the conventional nominal controllers, learning-based approaches have recently been introduced as alternatives owing to their robust performance against real-world uncertainties and model-free training in the simulation. In particular, the controllers learned via deep RL algorithms have been shown to perform effectively controlling conventional quadrotor~\cite{hwangbo2017control,lambert2019low,molchanov2019sim, penicka2022learning} and tilting hexarotor drones~\cite{lee2021low}. Despite the success in the experimental evaluations, those works still lack a sufficient stability guarantee to assure their feasibility in the real world.

\begin{figure*}[t!]
	\centering 
	\begin{subfigure}[b]{0.98\textwidth}
		\includegraphics[width=1.0\textwidth]{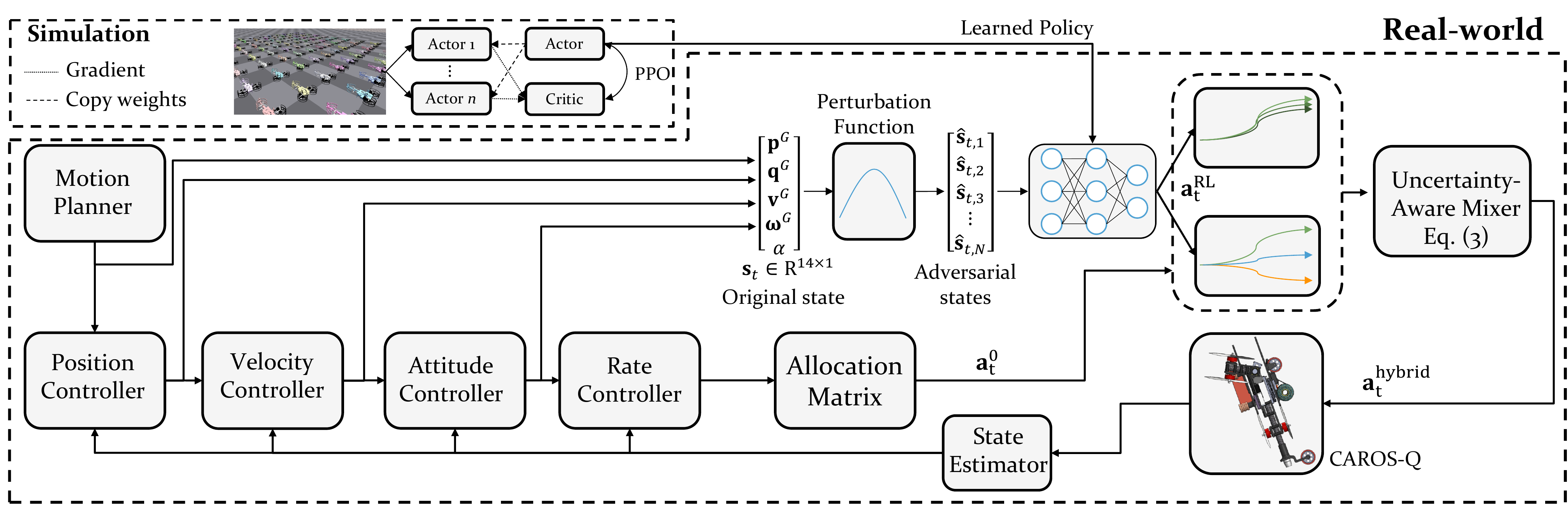}
	\end{subfigure}
	\captionsetup{font=footnotesize}
	\caption{Block diagram of Retro-RL applied to a drone platform, CAROS-Q~\cite{lee2021caros}. Definitions of the notations can be found in Sections~\ref{section:retro-rl} and~\ref{section:policy_learning}.}
	\label{figure:retro_rl_diagram_full}
\end{figure*}

Hybrid control frameworks have been proposed to overcome the limitation of both the nominal and learning-based control methods. The idea of hybrid control frameworks can be dated back to a simple Takagi-Sugeno fuzzy system applied as an intervention to a proportional-integral-derivative (PID) controller~\cite{gharieb2001fuzzy}. The modern approaches utilizing deep learning have focused on designing an adaptive model learned with neural networks~\cite{fujii1991neural,bauersfeld2021neurobem,bodie2019omnidirectional} or Gaussian Processes~\cite{torrente2021data} and deploying it with a nonlinear model-based controller. However, those methods required the model to be learned precisely and might fail in out-of-distribution conditions~\cite{levine2020offline}, i.e.~the condition when the test environment has a significant discrepancy with the training environment.

In~\cite{song2020learning,song2021policy}, a hybrid framework between a learning-based controller and a model predictive control (MPC), named high-MPC, was proposed. The main idea was to train the policy to tune the MPC's high-level parameters to achieve optimal performance. However, high-MPC assumes precise modeling of the MPC for a particular drone platform, and the main objective of the policy is only for optimizing the MPC parameters to perform aggressive drone maneuvers.

Closely related to our work is the hybrid framework between an RL-based control policy with a disturbance observer (DOB)~\cite{kim2019improving} or model reference adaptive control (MRAC)~\cite{guha2020mrac}. DOB or MRAC were used to reject disturbances such that the RL-based control policy can perform on a system without any disturbance. However, this framework relied heavily on the RL-based control policy as the main controller; thus, it might eventually fail in the out-of-distribution conditions.

In contrast with other works, we propose \emph{\textbf{Retro-RL}} (Reinforcing Nominal Controller with Deep RL), which boosts the robust performance of a nominal controller using a policy trained with an RL algorithm while preventing catastrophic failure due to out-of-distribution conditions of the neural network policy by exploiting a nominal controller when required. Retro-RL does not replace a nominal controller with a learning-based controller. Instead, both work in tandem, and each controller's contribution is determined statistically. The proposed Retro-RL is a generic framework; thus, it can be deployed on different robot platforms. This paper evaluates Retro-RL in controlling CAROS-Q, a tilting-rotor drone, for a wall-climbing task, extending our previous work~\cite{lee2021caros}.

In summary, the contributions of this paper are threefold.
\begin{itemize}
    	\item A novel control framework that fuses a model-based control method with a robust learning-based controller trained using a deep RL algorithm. The proposed framework prevents catastrophic failure of a neural network policy by querying the nominal controller when the neural network fails to provide a reasonable control action. The stability of the nominal (unperturbed) system's equilibrium in the proposed framework is proven via the Lyapunov stability theorem.
    	\item  A novel control mixing technique called uncertainty-aware control mixer is proposed. The proposed technique estimates the uncertainty of the learned policy for a given state and statistically weighs the inferred action to be mixed with the nominal controller's action.
    	\item  The proposed framework was validated through real-world experiments using CAROS-Q, including a wall-climbing scenario that might suffer from nonlinear aerodynamic effects due to the proximity of the wall.
    \end{itemize}

The remainder of this paper is organized as follows. Section~\ref{section:retro-rl} deeply explains the proposed method. Section~\ref{section:policy_learning} describes the algorithms for learning the control policy used in the Retro-RL framework. Section~\ref{section:experimental_results} presents experimental setups and results using a tilting hexarotor drone platform,  CAROS-Q. Finally, Section~\ref{section:conclusion} concludes the presented work and discusses potential future work.

\section{Retro-RL} \label{section:retro-rl}
This section presents a detailed description and analysis of the proposed Retro-RL. The implementation of Retro-RL to a tilting-rotor hexarotor, CAROS-Q~\cite{lee2021caros}, is shown in Fig.~\ref{figure:retro_rl_diagram_full}. Retro-RL is built on three main components, i.e.~a nominal controller, a robust performance auxiliary control policy, and an uncertainty-aware control mixer. The nominal controller is derived from~\cite{lee2021caros} using a reformulated dynamics model. The robust performance auxiliary control policy operates in tandem with the nominal controller and is trained using a distributed version of the on-policy deep RL algorithm to learn generalization via domain randomization.

\subsection{Preliminaries}
This paper assumes the environment as an infinite-horizon Markov decision process (MDP), defined by the tuple $\mathcal{M}=(\mathcal{S},\mathcal{A},d_0,p,r,\gamma)$. The state and action spaces defined by $\mathcal{\textbf{s}\in\mathcal{S}}$ and $\mathcal{\textbf{a}\in\mathcal{A}}$, respectively, are continuous; $d_0$ is the initial state distribution $d_0(\textbf{s}_0)$, $p$ is a state transition probability of the form $p(\textbf{s}_{t+1}|\textbf{s}_{t},\textbf{a}_{t})$ that describes how the system works (i.e., the dynamics model defined in a probabilistic manner), and $r:\mathcal{S}\times\mathcal{A}\to\mathcal{R}$ is a reward function.

The nominal and the auxiliary policy controllers are defined as $\pi_0$ and $\pi_\text{RL}$, respectively. $\pi_0$ is a deterministic controller, built on the nominal controller, and $\pi_\text{RL}$ is a probabilistic learning-based controller, trained with a deep RL algorithm.

\subsection{Uncertainty-Aware Control Mixer}
The control policies $\pi_0$ and $\pi_\text{RL}$ generate two different control actions
for a given state $\textbf{s}_t$. The uncertainty-aware control mixer is employed to combine the two control actions statistically. The main purpose of the uncertainty-aware control mixer is to apply uncertainty weight on the auxiliary control action generated by $\pi_\text{RL}$ using its degree of uncertainty, which is the estimation of uncertainty in the auxiliary policy's action for a given state (the implementation of the degree of uncertainty will be explained later in this section).. The mixer consists of three main steps: First, a batch of adversarial states $\hat{\textbf{s}}_t
\!=\!\begin{bmatrix}\hat{\textbf{s}}_{t,1}\!&\!\hat{\textbf{s}}_{t,2}\!&\!\dots\!&\!\hat{\textbf{s}}_{t,N}\end{bmatrix}$ with batch size $N$ is generated by corrupting $\textbf{s}_t$ with a perturbation function which follows the Gaussian distribution. Second, given the adversarial states, actions from the auxiliary control policy are sampled as follows:
\begin{equation}
\begin{split}
  &\hat{\textbf{a}}_t\sim\pi_\text{RL}(\hat{\textbf{a}}_t|\hat{\textbf{s}}_t)\\
  &\hat{\textbf{a}}_t
\!=\!\begin{bmatrix}\hat{\textbf{a}}_{t,1}\!&\!\hat{\textbf{a}}_{t,2}\!&\!\dots\!&\!\hat{\textbf{a}}_{t,N}\end{bmatrix}.
  \label{eqn:adversarial_action}
\end{split}
\end{equation}

Third, the degree of uncertainty of $\pi_\text{RL}$, is computed to weight the nominal and auxiliary control. Relative entropy theory is adopted to determine the degree of uncertainty, and the relative entropy is implemented using Kullback-Leibler divergence ($D_\text{KL}$)~\cite{kullback1951information}. 

The uncertainty measurement method is performed in a three-step process: First, the mean and variance of $\hat{\textbf{a}}_t$ are computed and stored as a Gaussian distribution $P$. Subsequently, any action, $\hat{\textbf{a}}_{t,i}$ is randomly sampled from $\hat{\textbf{a}}_t$ to construct a target distribution $Q$, which is a Gaussian distribution centered at $\hat{\textbf{a}}_{t,i}$ with variance $\sigma^2$, formulated as follows:
\begin{equation}
  Q\sim\mathcal{N}(\hat{\textbf{a}}_{t,i},\sigma^2),
  \label{eqn:target_distribution}
\end{equation}
where $\mathcal{N}(\cdot,\cdot)$ is the Gaussian distribution. The variance of $Q$ is a tunable hyperparameter that defines how much uncertainty from $\pi_{RL}$ is allowed. For a large variance, uncertain actions from $\pi_{RL}$ will be accepted by the uncertainty-aware mixer more often. In contrast, a small variance will reject uncertain actions more strictly. We chose a unit variance in this paper to be used in the experiments for mathematical simplicity. Finally, the uncertainty is measured by computing $D_\text{KL}(P\|Q)$.

Large $D_\text{KL}(P\|Q)$ value indicates that there is a large discrepancy between $P$ and $Q$. Then, because $Q$ is constructed from $\hat{\textbf{a}}_{t,i}$, the discrepancy between $P$ and $Q$ indicates that $\hat{\textbf{a}}_{t,i}$ cannot be represented by the Gaussian distribution formed by the adversarial actions, $\hat{\textbf{a}}_t$. Thus, it can be concluded that the $\pi_\text{RL}$ has a high uncertainty over the given state, $\textbf{s}_t$.

In contrast, small $D_\text{KL}(P\|Q)$ value demonstrates that $P$ and $Q$ are statistically similar, inferring that $\pi_\text{RL}$ has a low uncertainty over the given state, $\textbf{s}_t$. Finally, using the estimated degree of uncertainty, the hybrid control action can be formulated as follows:
\begin{equation}
  \textbf{a}_t^\text{hybrid}=D_\text{KL}(P\|Q)\textbf{a}_t^{0}+(1-D_\text{KL}(P\|Q))\textbf{a}_t^\text{RL},
  \label{eqn:hybrid_action}
\end{equation}

\noindent where $\textbf{a}_t^{0}$ is the control action drawn for the nominal control policy $\pi_0(\textbf{a}_t^{0}|\textbf{s}_t)$, and $\textbf{a}_t^\text{RL}$ is the control action drawn for the auxiliary control policy $\pi_0(\textbf{a}_t^\text{RL}|\textbf{s}_t)$. In~(\ref{eqn:hybrid_action}), it is implied that when $\pi_\text{RL}$ has a high uncertainty, its contribution to the control action decreases, and \textit{vice versa.}

\subsection{Quasi-Decoupling Controller} \label{sec:nominal_controller}
The main focus of this paper is applying Retro-RL on a tilting hexarotor drone built from our previous work~\cite{lee2021caros}, which has a Y-shaped coaxial configuration (Fig.~\ref{figure:caros-q}). The nominal control policy, $\pi_0$, is derived from the quasi-decoupling controller~\cite{lee2021caros}, which was selected over other control modalities including MPC, MRAC, and DOB thanks to the quasi-decoupling controller's simplicity and not requiring any online optimization during deployment~\cite{lee2021caros}.

In the previous work, the drone's body frame was defined at the center of gravity (CoG). Hence, the body frame location is no longer valid when the CoG changes owing to the arm-tilting maneuver. In this paper, the body frame is redefined and fixed on the tilting axle, resulting in a new formulation of the thrusts and moments equilibrium equation as follows:
\begin{figure}[t!]
	\centering 
	\begin{subfigure}[b]{0.48\textwidth}
		\includegraphics[width=1.0\textwidth]{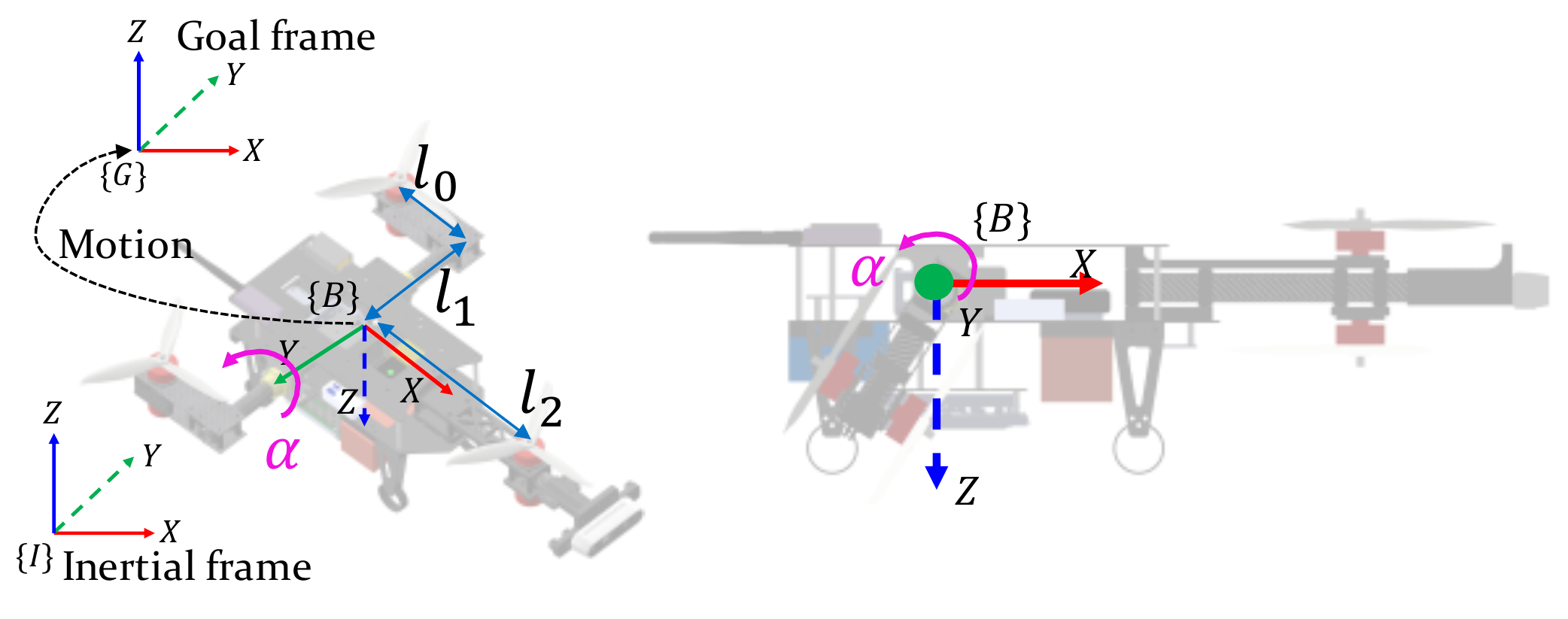}
	\end{subfigure}
	\captionsetup{font=footnotesize}
	\caption{Hardware configuration of the CAROS-Q used in this work with some modifications from~\cite{lee2021caros}. $\{I\}$, $\{B\}$, and $\{G\}$ correspond to the inertial, body, and goal frame, respectively. Details of the hardware parameters can be found in Section~\ref{sec:nominal_controller} and Table~\ref{table:hardware_parameters}.}
	\label{figure:caros-q}
\end{figure}
\begin{equation}
\begin{split}
  &F_x^B:F_1\sin{\alpha}=mg\sin{\theta_\text{des}}\\
  &F_z^B:F_2\cos{\alpha}+F_2=mg\cos{\theta_\text{des}}\\
  &M_y^B:-F_1l_0+F_2l_2-mg(z_\text{CoG}\sin{\theta}-x_\text{CoG}\sin{\theta})=0,
  \label{eqn:nonlinear_equations}
\end{split}
\end{equation}
where $F_x^B$ and $F_z^B$ are the thrusts in the $x$ and $z$ axis of the body frame, respectively, $M_y^B$ is the moment along the $y$ axis of the body frame, and $F_1$ and $F_2$ are the thrusts generated by the front and rear rotors, respectively. The CoG terms, $z_\text{CoG}$ and $x_\text{CoG}$ are introduced owing to the definition of the body frame on the tilting axle. At the implementation level, $z_\text{CoG}$ and $x_\text{CoG}$ are estimated using an equation obtained from the CAD model data.

$\theta_\text{des}$ is the body frame's desired pitch angle that allows pitch change while maintaining the current position. It can be obtained by solving~(\ref{eqn:nonlinear_equations}),  where $\theta$ is the current pitch angle of the body frame, and $\alpha$ is the front rotors' tilting angle w.r.t. the body frame. The hardware-related parameters, $l_0$, $l_2$, $x_\text{CoG}$, $z_\text{CoG}$, $m$, and $g$ are the rotor offset length, distance from the body frame to rear rotors, $x$ and $z$ position of the CoG, the mass of the drone, and gravitational acceleration, respectively.

The resulting $\theta_\text{des}$ is used as an input for the attitude controller as the desired orientation in quaternion form, $\textbf{q}_\text{des}$. The subsequent steps, including rate controller and thrusts allocation, follow the standard procedures for multirotor control in the PX4 control stack. In this paper, instead of fully learning the $\theta_\text{des}$ and $\alpha$ combinations from scratch~\cite{lee2021low}, the quasi-decoupling control is adopted in the reward function to guide the policy to discover the optimal $\theta_\text{des}$. It was implemented by setting $\theta_\text{des}$ as the target pitch in the orientation reward.

\section{Robust Policy Learning} \label{section:policy_learning}
\subsection{State Representation}
This paper assumes the full observability of the multirotor's state. Similar to ~\cite{lee2021caros,hwangbo2017control}, the multirotor's pose is represented in the goal frame to control the drone by simply changing the location of the goal frame in the inertial frame. The observation is a $14\times1$ vector $\begin{bmatrix}\textbf{\text{p}}^G & \textbf{\text{q}}^G & \textbf{\text{v}}^G & \boldsymbol{\omega}^G & \alpha\end{bmatrix}^T$, where $\textbf{\text{p}}^G$, $\textbf{\text{q}}^G$, $\textbf{\text{v}}^G$, and $\boldsymbol{\omega}^G$ are the drone's position, attitude quaternion, linear velocity, and angular velocity, respectively, represented in the goal frame. As mentioned in Section~\ref{sec:nominal_controller}, the quasi-decoupling outputs $\theta_\text{des}$ as the goal pitch angle for the corresponding $\alpha$. In this work, $\theta_\text{des}$ is adopted as the goal pitch in $\textbf{\text{q}}^G$. 
The policy network receives the observation vector and outputs a $6\times1$ action vector consisting of six individual rotors' thrust. In the real-world implementation, the thrusts are converted into rotor speeds using thrust mapping as in~\cite{kamel2018voliro}.

\subsection{Provably Stable Reward Function}
Reward design is a crucial aspect of deep RL because it shapes the behavior of the learned policy. Moreover, its importance increases for a system with a high stability requirement. Other works on multirotor drone control~\cite{hwangbo2017control} constructed the reward function by considering pose and velocity errors and including a power consumption term to learn a pose-tracking policy with a minimum control effort~\cite{lee2021low}. However, previous works lacked the stability guarantee of the drone. This paper shows that the choice of pose and velocity tracking error as a part of the reward function can guarantee the nominal system's (without any perturbation) stability when incorporated into a hybrid setting proposed in Retro-RL, which is proved using the Lyapunov stability theorem in this section.

\begin{theorem} \label{theorem:stability}
Let a tilting hexarotor drone system be described with a state and system dynamics as:
\begin{equation}
\begin{split}
  &\textbf{\text{x}}=\begin{bmatrix}\textbf{\text{p}}^G & \textbf{\text{q}}^G & \textbf{\text{v}}^G & \boldsymbol{\omega}^G & \alpha\end{bmatrix}^T\\
  &\dot{\textbf{\text{x}}}=f(\textbf{\text{x}}),
  \label{eqn:system_dynamics}
\end{split}
\end{equation}

\noindent and its equilibrium point is defined in the origin. For the given system, there exists a candidate Lyapunov function $V(\textbf{\text{x}})$, which is a radially unbounded positive definite function and its first derivative, $\dot{V}(\textbf{\text{x}})$ is negative definite. Hence, the equilibrium point is globally, uniformly, and asymptotically stable.
\end{theorem}

\begin{proof}
To prove the stability of the equilibrium point, we define the appropriate Lyapunov function that is radially unbounded and positive definite, whose first derivative is negative definite. Let a reward function be defined as follows:
\begin{equation}
\begin{split}
  &r(\textbf{\text{x}},t)=r_{\textbf{\text{p}}}(\textbf{\text{p}}^G,t)+r_{\textbf{\text{q}}}(\textbf{\text{q}}^G,t)+r_{\textbf{\text{v}}}(\textbf{\text{v}}^G,t)+r_{\boldsymbol{\omega}}(\boldsymbol{\omega}^G,t),\\
  &r_{\textbf{\text{k}}}(\textbf{\text{k}},t)=\frac{1}{1+\|\textbf{\text{k}}\|};\textbf{\text{k}}\in(\textbf{\text{p}}^G,\textbf{\text{q}}^G,\textbf{\text{v}}^G,\boldsymbol{\omega}^G),
  \label{eqn:reward_function}
\end{split}
\end{equation}
where $r_{\textbf{\text{p}}}(\textbf{\text{p}}^G,t)$, $r_{\textbf{\text{q}}}(\textbf{\text{q}}^G,t)$, $r_{\textbf{\text{v}}}(\textbf{\text{v}}^G,t)$, and $r_{\boldsymbol{\omega}}(\boldsymbol{\omega}^G,t)$ are the position, attitude, linear velocity and angular velocity rewards, respectively. All these reward functions follow the form of $r_{\textbf{\text{k}}}(\textbf{\text{k}},t)$, which is a positive definite function.

As an example of the proof, let a candidate Lyapunov function for the position reward be defined as follows:
\begin{equation}
    V_{\textbf{\text{p}}}(\textbf{\text{p}}^G,t)=1+\|\textbf{\text{p}}^G\|,
    \label{eqn:lyapunov_position}
\end{equation}
which is a positive definite function and radially unbounded. The first derivative of this candidate Lyapunov function can be derived as follows:
\begin{equation}
    \dot{V}_{\textbf{\text{p}}}(\textbf{\text{p}}^G,t)=\pdv{V_\textbf{p}}{t}+\pdv{V_\textbf{p}}{\textbf{p}^G}f(\textbf{p}^G,t)=\frac{x\dot{x}+y\dot{y}+z\dot{z}}{\sqrt{x^2+y^2+z^2}},
    \label{eqn:dot_lyapunov_position}
\end{equation}
where $x$, $y$, and $z$ are the 3D position coordinates of the drone's body frame relative to the goal frame, and $\dot{x}$, $\dot{y}$, and $\dot{z}$ are their first derivatives w.r.t. time.

In its original form, the negative definiteness of ~(\ref{eqn:dot_lyapunov_position}) cannot be concluded directly. However, without loss of generality, this problem can be solved by enforcing any feedback control law on the position controller, which is made possible from the hybrid structure of Retro-RL using the feedback control law as follows:
\begin{equation}
    \dot{\textbf{\text{p}}}^G=-\textbf{\text{K}}\textbf{\text{p}}^G,
    \label{eqn:feedback_position}
\end{equation}
where the control gain $\textbf{\text{K}}$ is a positive diagonal matrix, making $\dot{V}_{\textbf{\text{p}}}(\textbf{\text{p}}^G,t)$ be negative definite. Therefore, the negative definiteness of $\dot{V}_{\textbf{\text{p}}}(\textbf{\text{p}}^G,t)$ is guaranteed. The Retro-RL algorithm enforces the stability by assigning pose and velocity goals generated by the nominal controller. These goals are used as references and the tracking errors construct the input vector for the policy network. Hence, adopting the reward function in the form of ~(\ref{eqn:reward_function}) in Retro-RL results in a globally, uniformly, and asymptotically stable equilibrium point. 
\end{proof}

Because all the other terms of the reward function follow the form of $r_{\textbf{\text{k}}}(\textbf{\text{k}},t)$, their stability can be proven using the same approach. Note that, in~(\ref{eqn:feedback_position}), the feedback control law is given as a simple proportional gain, $\textbf{K}$, for simplicity. However, any class of negative feedback control law, such as PID (used in our position controller) and linear quadratic regulator (LQR) can theoretically work with the proposed framework.

Moreover, the stability analysis of the perturbed system can be extended by introducing an additional perturbation, $g(\textbf{x},t)$, to $\dot{\textbf{x}}$ in~(\ref{eqn:system_dynamics}). For the sake of generality, $g(\textbf{x},t)$ is assumed to be a non-vanishing perturbation. The candidate Lyapunov function's first derivative for the perturbed system is derived as follows:
\begin{equation}
    \dot{V}^\epsilon_{\textbf{\text{p}}}(\textbf{\text{p}}^G,t)=\pdv{V_\textbf{p}}{t}+\pdv{V_\textbf{p}}{\textbf{p}^G}f(\textbf{p}^G,t)+\pdv{V_\textbf{p}}{\textbf{p}^G}g(\textbf{p}^G,t).
    \label{eqn:perturbed_lyapunov_first_derivative}
\end{equation} 
According to Theorem~\ref{theorem:stability}, the first two terms of the right-hand side of~(\ref{eqn:perturbed_lyapunov_first_derivative}) are equal to~(\ref{eqn:dot_lyapunov_position}) and hence, negative definite. Substituting~(\ref{eqn:dot_lyapunov_position}) into~(\ref{eqn:perturbed_lyapunov_first_derivative}) yields:
\begin{equation}
\begin{split}
    \dot{V}^\epsilon_{\textbf{\text{p}}}(\textbf{\text{p}}^G,t)&=V_{\textbf{\text{p}}}(\textbf{\text{p}}^G,t) + \frac{x+y+z}{\sqrt{x^2+y^2+z^2}}g(\textbf{p}^G,t)\\
    &\leq V_{\textbf{\text{p}}}(\textbf{\text{p}}^G,t) + \abs{\frac{x+y+z}{\sqrt{x^2+y^2+z^2}}}\norm{g(\textbf{p}^G,t)}.
    \label{eqn:lyapunov_perturbed}
\end{split}
\end{equation}

To guarantee the negative definiteness of $\dot{V}^\epsilon_{\textbf{\text{p}}}(\textbf{\text{p}}^G,t)$, we can bound the second term of the right-hand side of (\ref{eqn:lyapunov_perturbed}) to be small enough, such that $\dot{V}^\epsilon_{\textbf{\text{p}}}(\textbf{\text{p}}^G,t)$ is still more negative than the last term of~(\ref{eqn:lyapunov_perturbed}). To do that, one can bound $\norm{g(\textbf{p}^G,t)}$ under some small value, which can be determined with further assumption or knowledge on the perturbation's properties.

Although the last term of (\ref{eqn:lyapunov_perturbed}) can be bounded even with very small values of $(x,y,z)$, it has different limits when approaching $0$ from different directions ($0^+$ or $0^-$). Hence, further assumption on the perturbation's bound, $\norm{g(\textbf{p}^G,t)}$ is required to simplify the equation and obtain a distinct bound to show the perturbed system's ultimate boundedness. However, we leave further investigation on the perturbation's type and its detailed properties for our future work.

\subsection{Domain Randomization}
\begin{table}[t!]
\centering
\captionsetup{font=footnotesize, justification=centering}
\caption{Hardware Parameters}
\label{table:hardware_parameters}
\begin{center}
\begin{tabular}{lccc}
\hline\hline
Parameter                 & Value           & Randomization range & Units \\\hline
Weight                    & $3.475$  & $[3.2,3.6]$ & $\mathrm{~kg}$     \\ 
Max. thrust of each rotor~\cite{rotors} & $19.94$   & - & $\mathrm{~N}$    \\ 
Stall torque of servo~\cite{robotis}         & $10.6$  & - & $\mathrm{~Nm}$   \\ 
$l_0$                     & $85.55$  & $[80.0,90.0]$ & $\mathrm{~mm}$  \\ 
$l_1$                     & $182.0$  & $[170.0,190.0]$ & $\mathrm{~mm}$    \\ 
$l_2$                     & $287.0$  & $[270.0,300.0]$ & $\mathrm{~mm}$   \\ 
Nominal CoG $(x)$     & $55.2$      & $[40.0,65.0]$ & $\mathrm{~mm}$    \\
Nominal CoG $(y)$     & $0.0$      &$[-10.0,10.0$ & $\mathrm{~mm}$    \\
Nominal CoG $(z)$     & $8.2$      & $[3.0,15.0]$ & $\mathrm{~mm}$   \\
\hline\hline
\end{tabular}
\end{center}
\vspace{-0.3cm}
\end{table}

Domain randomization enables a policy trained in the simulation to learn a broad spectrum of uncertainties, enhancing the policy's robustness when deployed in the real-world environment. Thus, this work exploits domain randomization to capture parametric hardware uncertainties into the learned policy. Domain randomization is applied with a fixed frequency to every resetting environment in the simulation. Unlike in~\cite{lee2021low}, where the policy was trained to learn a hovering and pose-changing combined task, the policy is trained only for a hovering task here. Subsequently, the agent perceives the tilting angle, $\alpha$, as a variable in the environment. It is realized by including the pose-changing task in the domain randomization. $\alpha$ is set to a value randomly selected from the set $[0^\circ,110^\circ]$ for each domain randomization. Therefore, the policy can further explore the hovering state for any given $\alpha$. Moreover, only half of the environments are randomized to preserve the nominal hovering capability, while the other half continues learning how to hover at $\alpha=0^\circ$.

Domain randomization was also applied to other parameters, i.e.~mass, nominal CoG location, and distance from the body frame to the rotors. These parameters were randomized since the dynamics of CAROS-Q non-linearly depend on them. Therefore, the policy was trained to learn these parametric uncertainties. A summary of the randomization range of these parameters is presented in Table~\ref{table:hardware_parameters}, where the nominal values were measured from the actual drone's components and 3D CAD (Computer-Aided Drawing) software.

\begin{table}[t!]
\centering
\captionsetup{font=footnotesize, justification=centering}
\caption{Simulation Parameters}
\label{table:simulation_parameters}
\begin{center}
\begin{tabular}{lccc}
\hline\hline
Parameter                 & Value       \\\hline
Number of steps per episode                   & $500$   \\ 
Number of actors & $8,\!192$  \\ 
Domain randomization freq.                   & $1,\!000$ \\ 
Activation function & ReLU~\cite{agarap2018deep}\\
Learning rate & $0.001$\\ 
Clipping range & $0.2$\\
Optimizer & Adam~\cite{kingma2014adam}\\
Discount factor &$0.99$\\
GAE factor & $0.95$\\

\hline\hline
\end{tabular}
\end{center}
\vspace{-0.3cm}
\end{table}

\subsection{Algorithm for Policy Learning} \label{section:algorithm}
The auxiliary control policy, $\pi_\text{RL}$, is designed to provide robust performance to support the nominal controller. Thus, $\pi_\text{RL}$ is trained in the simulation with domain randomization~\cite{tobin2017domain}, resulting in generalization for a  broad spectrum of uncertainties. Learning a policy with domain randomization can be conducted via several methods, for instance, meta-learning, multi-task learning, or parallel exploration. In this paper, a fast adaptation ability from meta-learning does not become the aim since it requires gradient updates online, which will cause a less favorable computational burden to the platform. Moreover, considering that the proposed system randomizes only several parametric uncertainties in the drone and not several different tasks, multi-task learning is not appropriate either. Therefore, a parallel exploration method was chosen in this work.

The policy learning used in this work is based on the synchronous version of~\cite{mnih2016asynchronous}, called the advantage actor-critic (A2C). Additionally, the continuous actor-critic architecture is utilized, and the policy is optimized using the proximal policy optimization (PPO) algorithm~\cite{schulman2017proximal}. 

\section{Experimental Results} \label{section:experimental_results}
\subsection{Simulation Setup}
\begin{figure}[t!]
	\centering 
	\begin{subfigure}[b]{0.48\textwidth}
		\includegraphics[width=1.0\textwidth]{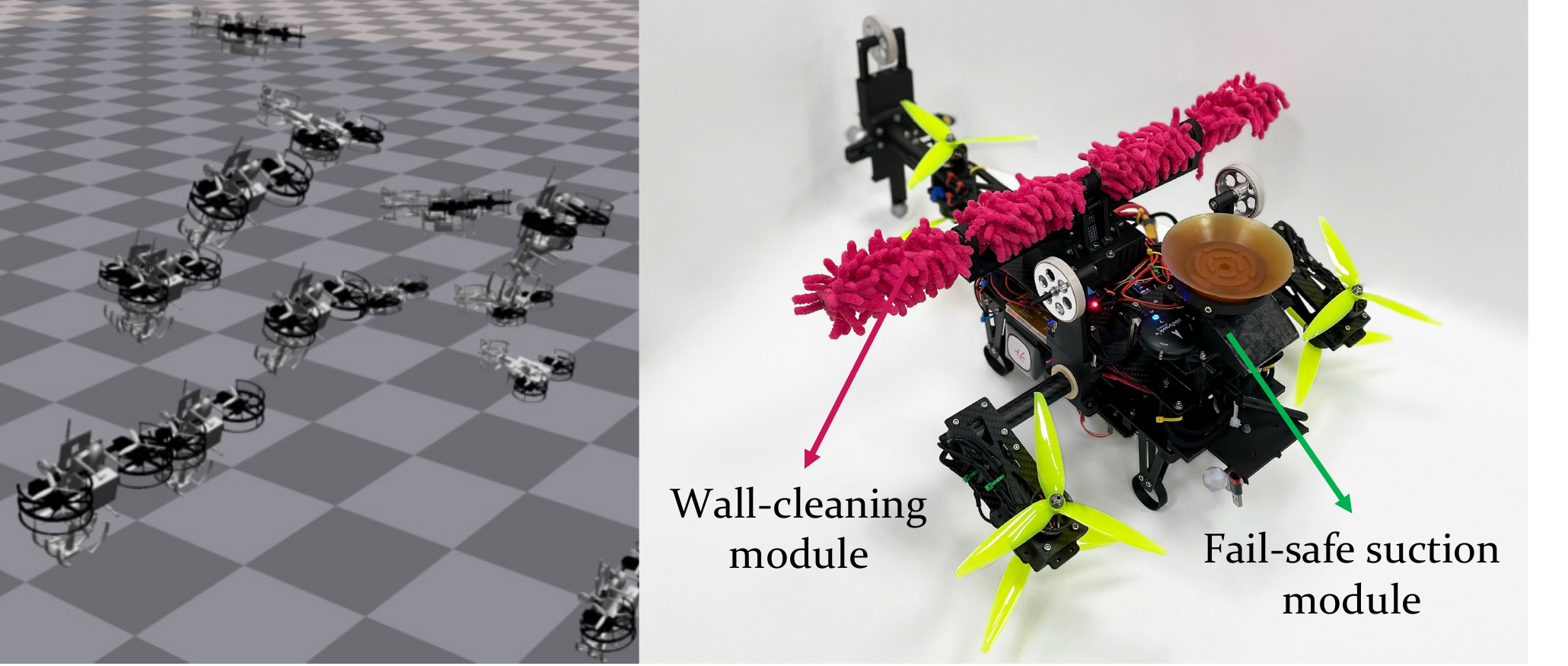}
	\end{subfigure}
	\captionsetup{font=footnotesize}
	\caption{NVIDIA Isaac Gym simulation (left) and real-world platform (right) of CAROS-Q. Note that even though there are discrepancies between the model trained in simulation and the real-world platform due to additional wall-cleaning and fail-safe modules as shown in the figure, the proposed Retro-RL successfully controls the CAROS-Q with satisfactory performance.}
	\label{figure:isaacgym}
\end{figure}

NVIDIA Isaac Gym~\cite{makoviychuk2021isaac}, a GPU-accelerated physics simulator, was selected as the simulation environment for its capability to handle multiple simulations rapidly. The distributed RL algorithms can be implemented using this simulator, and its performance has been benchmarked on different robotics platforms in other studies~\cite{makoviychuk2021isaac,rudin2021learning}. The simulation was run on a desktop PC with an Intel Core i7-8700 CPU @ 3.20 GHz, 32 GB RAM, and an NVIDIA RTX 3060Ti GPU. The constructed environment for training the policy is shown in Fig.~\ref{figure:isaacgym}. Each episode of the simulation lasts for 500 environment steps, where each step equals to $0.01s$ in real-world time. The policy network was trained using the algorithm specified in Section~\ref{section:algorithm}. The network is a feedforward network with two hidden layers, which have 128 neurons/layer. The learning parameters are detailed in Table~\ref{table:simulation_parameters}.

\subsection{Simulation Results}
\begin{figure}[t!]
	\centering 
	\begin{subfigure}[b]{0.48\textwidth}
		\includegraphics[width=1.0\textwidth]{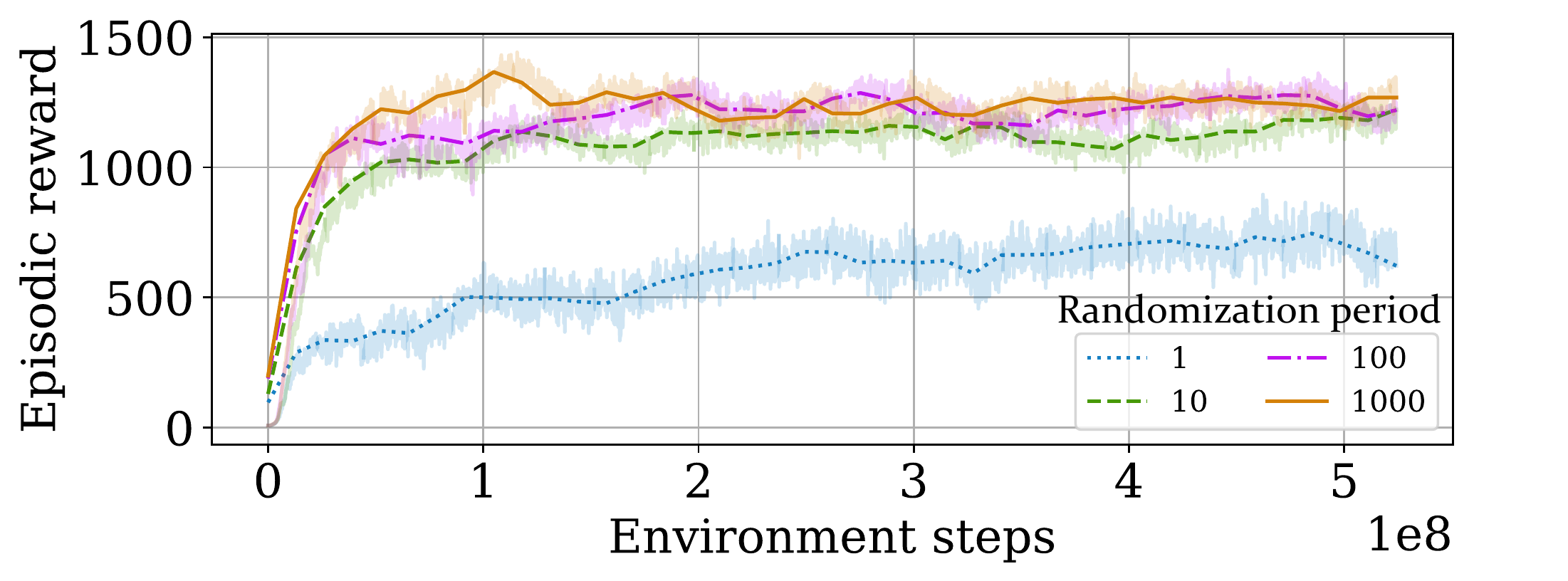}
	\end{subfigure}
	\captionsetup{font=footnotesize}
	\caption{Learning curves of the policy trained with PPO~\cite{schulman2017proximal} for four different domain randomization periods. The randomization period is equal to the number of environment steps taken in simulation before randomizing the drone parameters.}
	\label{figure:learning_curves}
\end{figure}

The simulations were performed to train an auxiliary policy, $\pi_\text{RL}$, to fly the drone in a stable hovering state. Moreover, the effect of different randomization periods on the optimal policy's performance was evaluated. As shown in Fig.~\ref{figure:learning_curves}, higher randomization periods resulted in a better policy, indicated by the higher episodic reward, and a more robust performance, indicated by the variance of the cumulative reward. Thanks to the massive parallel simulation of the NVIDIA Isaac Gym, the agent could explore 500 million environment steps within 20 minutes using a single NVIDIA RTX 3060Ti GPU.

\begin{figure}[t!]
	\centering 
	\begin{subfigure}[b]{0.48\textwidth}
		\includegraphics[width=1.0\textwidth]{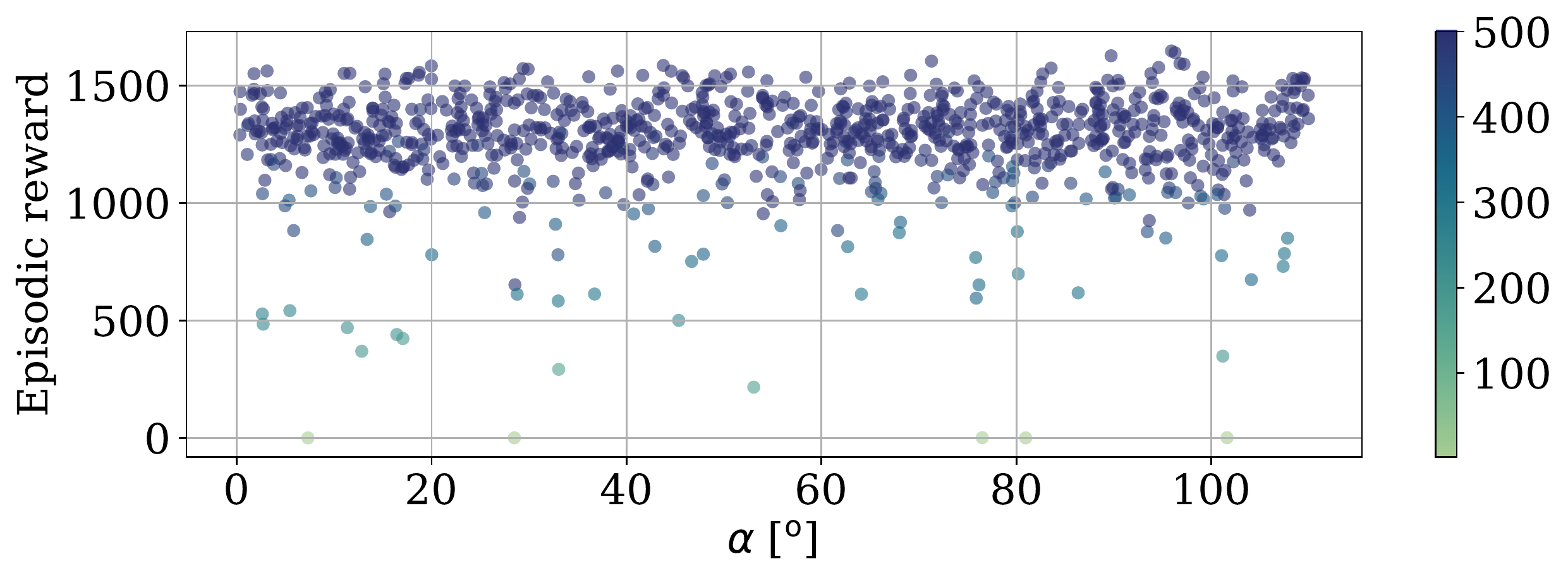}
	\end{subfigure}
	\captionsetup{font=footnotesize}
	\caption{Reward distribution of the policy on the evaluation stage. The evaluation was conducted by running the policy to control the CAROS platform to hover at a certain tilting angle ($\alpha$) for $500$ environment steps.}
	\label{figure:reward_distribution}
\end{figure}

Furthermore, the effectiveness of the domain randomization formulation for separating hovering and pose-changing tasks was validated, as shown in Fig.~\ref{figure:reward_distribution}. The figure shows reward distribution over $1,\!000$ evaluations, with 500 steps for each evaluation. The learned policy successfully flew the drone to hover at the desired position with a high cumulative reward and precision, indicated by the reward distribution over $\alpha$. The drone occasionally failed to hover; however, this was due to the extreme initial pose of the drone, for instance, when the roll angle of the drone was more than $30^\circ$. However, that condition can be avoided in the real-world implementation because the drone never starts to fly with an extreme initial state.

\subsection{Real-World Experimental Setup}
The experiments were carried out using the CAROS-Q with a slightly larger structure and a wall-cleaning module on its top plate. Its parameters are detailed in Table~\ref{table:hardware_parameters}. The actuators of the platform included six 7-inch propellers, six rotors, and a servo motor. A Pixhawk4 Mini was used as an onboard low-level controller for interfacing with the rotors. A Jetson Xavier NX board was used to run the high-level controller, responsible for generating waypoints, communicating with the OptiTrack Prime$^\text{X}$ 13 motion capture system to obtain the platform's pose estimate, and running the Retro-RL policy at $30$ Hz rate. 

The control performance of Retro-RL was compared with other state-of-the-art controllers which can be adopted to CAROS-Q, i.e. (1) \textbf{Quasi-decoupling}, the nominal controller; and (2) \textbf{PPO adopted}, which is an RL-based policy trained with PPO as the vanilla deep RL algorithm and deployed on the CAROS-Q platform without any hybridization with the nominal controller. For fair comparisons, the quasi-decoupling controller has been appropriately tuned to achieve satisfactory performance and was also used as the nominal controller in Retro-RL. Furthermore, the neural network policies in PPO adopted and Retro-RL were only trained in the simulator and zero-shot transferred to the real world. The comparative experiments consisted of pose tracking and trajectory tracking. Finally, the Retro-RL controller was tested for a wall-climbing maneuver using the CAROS-Q.

\subsection{Real-World Experimental Results}
\begin{figure}[t!]
	\centering 
	\begin{subfigure}[b]{0.48\textwidth}
		\includegraphics[width=1.0\textwidth]{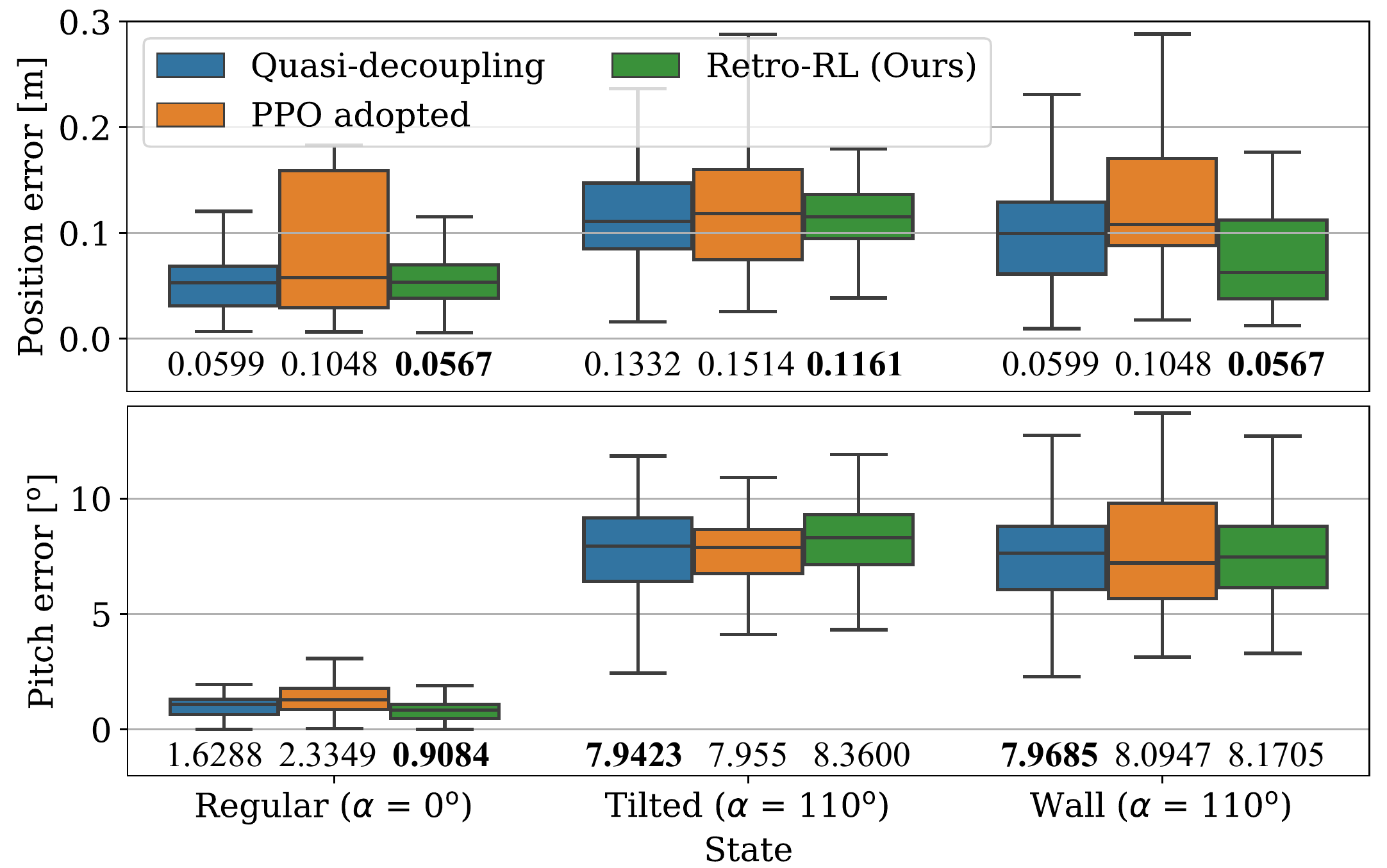}
	\end{subfigure}
	\captionsetup{font=footnotesize}
	\caption{Pose tracking comparison. The first box plot shows the RMSE (Root Mean Squared Error) for position tracking and the second plot shows the absolute pitch error. The measurements were taken over $15$ seconds of hovering. Values below the box plots are RMSEs over time of the corresponding flight mode.}
	\label{figure:pose_tracking_comparison}
	\vspace{-0.2cm}
\end{figure}

\subsubsection{\textbf{4-DoF Pose Following}}
The pose following accuracy of the proposed controller was tested on three different hovering modes, namely, regular, tilted, and near-the-wall hovering modes, which correspond to the hovering state without any tilting maneuver ($\theta=0^{\circ}$), obstacle-free hovering with $\theta_\text{des}=90^{\circ}$, and hovering in the proximity of the wall with $\theta_\text{des}=90^{\circ}$, respectively. As noted in~\cite{lee2021caros}, the CAROS-Q might suffer from a combination of wall and ceiling effects owing to its nonlinear aerodynamic effects generated by the proximity to the wall~\cite{davis2018aerodynamic,kocer2018centralized}. 

The pose following errors are visualized in Fig.~\ref{figure:pose_tracking_comparison}, with the RMSE values attached below each box plot. Additionally, we validated the improvement of Retro-RL against the baselines using $\textit{p}$-value of the two-sample \textit{t}-Test with $5\%$ significance level. The results are reported in Table~\ref{table:ttest}.

From the result in Fig.~\ref{figure:pose_tracking_comparison}, it is evident that the proposed Retro-RL outperformed the baseline controllers in the 3-DoF position following. The improvement is also validated within the $5\%$ significance level. However, the pitch following was not outperforming other baseline controllers, although the error is still within an acceptable threshold. Our argument for this phenomenon is that the Retro-RL controller sacrifices pitch following performance for minimum position following error. This could happen because the auxiliary policy was trained mainly to hover, and there are other DoFs other than pitch that need to be followed. To improve the pitch following performance, it is better to actively control $\alpha$, unlike the quasi-decoupling method that was not actively controlling $\alpha$. However, we leave this as a direction for future work.

\begin{table}[]
\centering
\captionsetup{font=footnotesize, justification=centering}
\caption{\textit{p}-value of two-Sample \textit{t}-Test between Algorithms. Bold values indicate \textit{p}-value below $5\%$. (\textbf{RQ}: Retro-RL vs Quasi-decoupling, \textbf{RP}: Retro-RL vs PPO adopted)}
\label{table:ttest}
\begin{tabular}{lcccccc}\toprule
State          & \multicolumn{2}{c}{Regular} & \multicolumn{2}{c}{Tilted} & \multicolumn{2}{c}{Wall} \\
Sample     & RQ           & RP           & RQ           & RP          & RQ          & RP         \\
\midrule
Pos. error & $0.151$             & $\boldsymbol{10^{-3}}$             & $\boldsymbol{0.044}$             & $\boldsymbol{10^{-3}}$            & $\boldsymbol{10^{-4}}$            & $\boldsymbol{10^{-4}}$           \\
Pit. error    & $\boldsymbol{0.002}$             & $\boldsymbol{0.1}$             & $\boldsymbol{0.011}$             & $\boldsymbol{0.025}$            & $0.394$            & $0.634$          \\
\bottomrule
\end{tabular}
\vspace{-0.3cm}
\end{table}

\subsubsection{\textbf{Trajectory Tracking}}
The pose tracking performance was evaluated further by tracking a lemniscate trajectory when the drone {was} fully-tilted. This experiment was carried out to observe the superiority of the proposed algorithm in controlling the drone in a high-maneuverability task on a challenging drone configuration. 

\begin{figure}[t!]
	\centering 
	\begin{subfigure}[b]{0.48\textwidth}
		\includegraphics[width=1.0\textwidth]{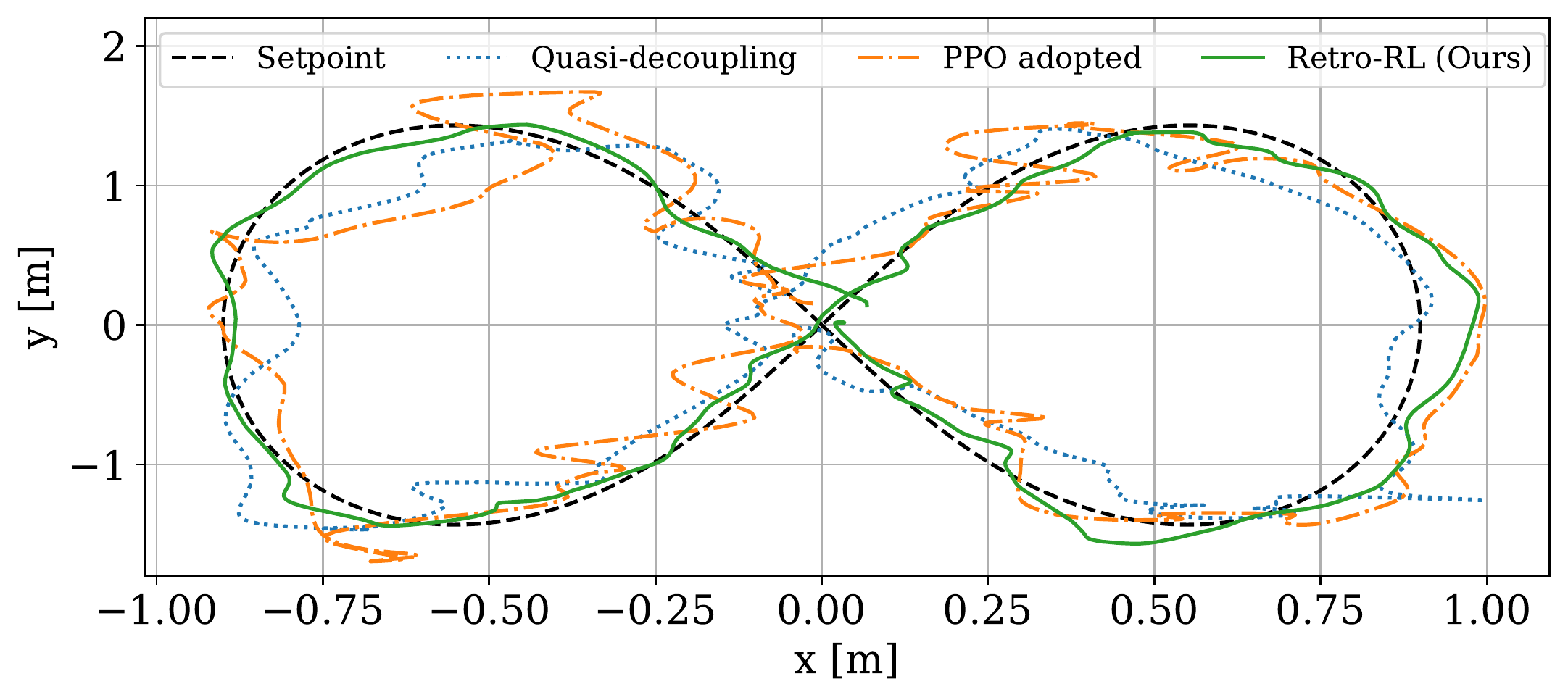}
	\end{subfigure}
	\captionsetup{font=footnotesize}
	\caption{Lemniscate trajectory tracking comparison of the CAROS-Q controlled with different controllers for $\alpha = 110^{\circ}$. The space used for experiment, along with its inertial frame definition can be seen in Fig.~\ref{figure:wall_climbing}.}
	\label{figure:trajectory_tracking}
\end{figure}

The evaluation result is presented qualitatively in Fig.~\ref{figure:trajectory_tracking}. The drone controlled using Retro-RL successfully navigated through the lemniscate trajectory and outperformed other controllers, thanks to its guaranteed stability and the robust performance from the learned policy. Most notably, compared with the PPO-adopted controller, the Retro-RL controller was more stable because of the activation of the uncertainty-awareness of the controller.

\begin{figure}[t!]
	\centering 
	\begin{subfigure}[b]{0.478\textwidth}
		\includegraphics[width=1.0\textwidth]{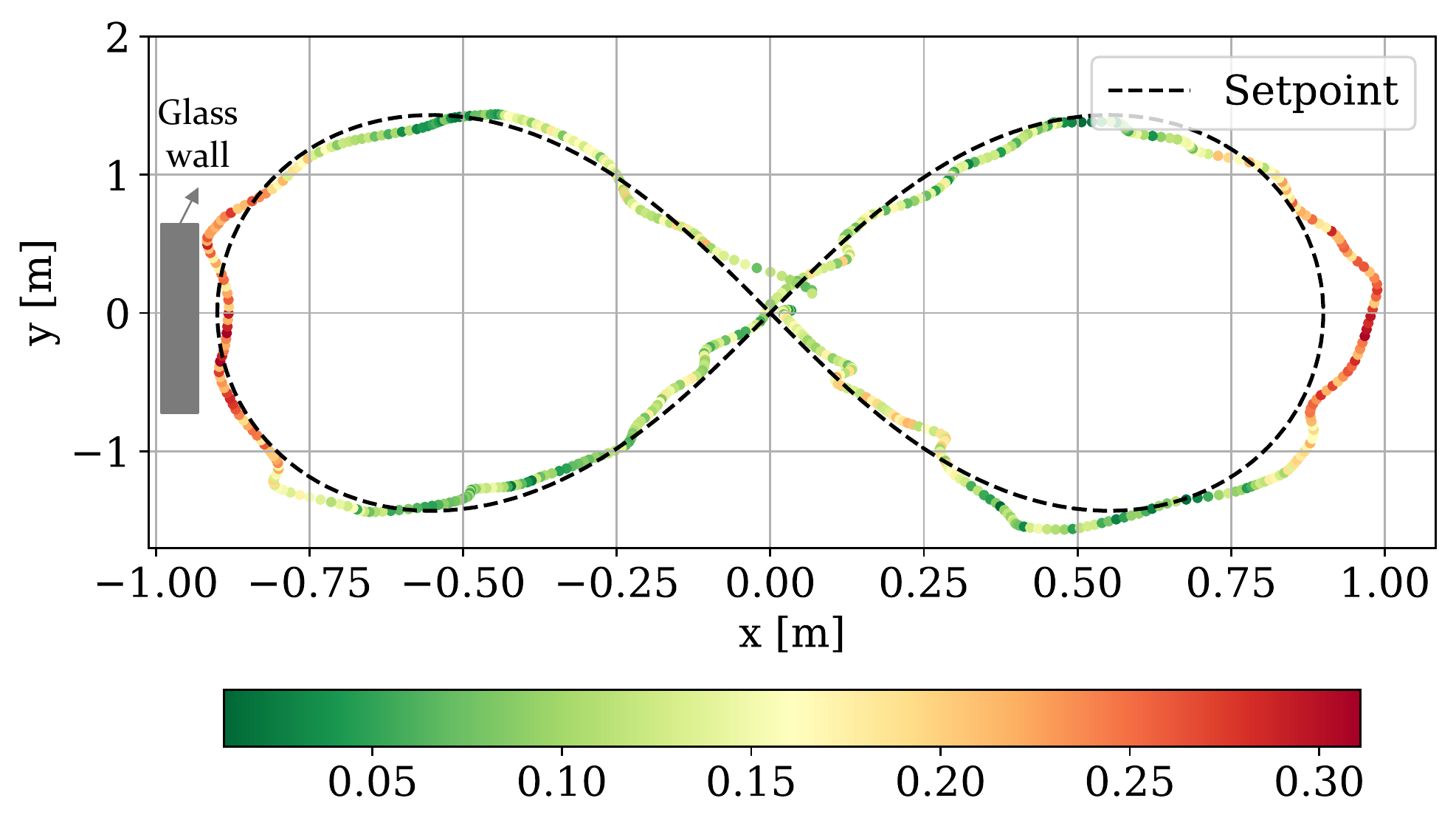}
	\end{subfigure}
	\captionsetup{font=footnotesize}
	\caption{Uncertainty weight during the lemniscate trajectory tracking using Retro-RL. The colorbar indicates the normalized $D_\text{KL}(P|Q)\in[0,1]$.}
	\label{figure:entropy_weighting}
\end{figure}

Fig.~\ref{figure:entropy_weighting} visualizes the uncertainty weight in terms of $D_\text{KL}(P\|Q)$ from the auxiliary policy used in the experiment of Fig.~\ref{figure:trajectory_tracking}. The uncertainty weight dynamically changed during the flight depending on the condition of the trajectory. The uncertainty weight mostly increased around sharp curves, indicating that the auxiliary policy is unreliable and must be supported further by the nominal controller. The entropy weight tended to be high near $x\!=\!-1$ or $x\!=\!1$ because of the sideways motion that is hard to be performed by CAROS-Q in its tilted mode, even without the existence of the wall. Moreover, the wall at $x\!=\!-1$ also produced additional disturbance due to the wall effect.


\begin{figure}[t!]
    \centering
    \begin{subfigure}[b]{0.49\textwidth}
		\includegraphics[width=1.0\textwidth]{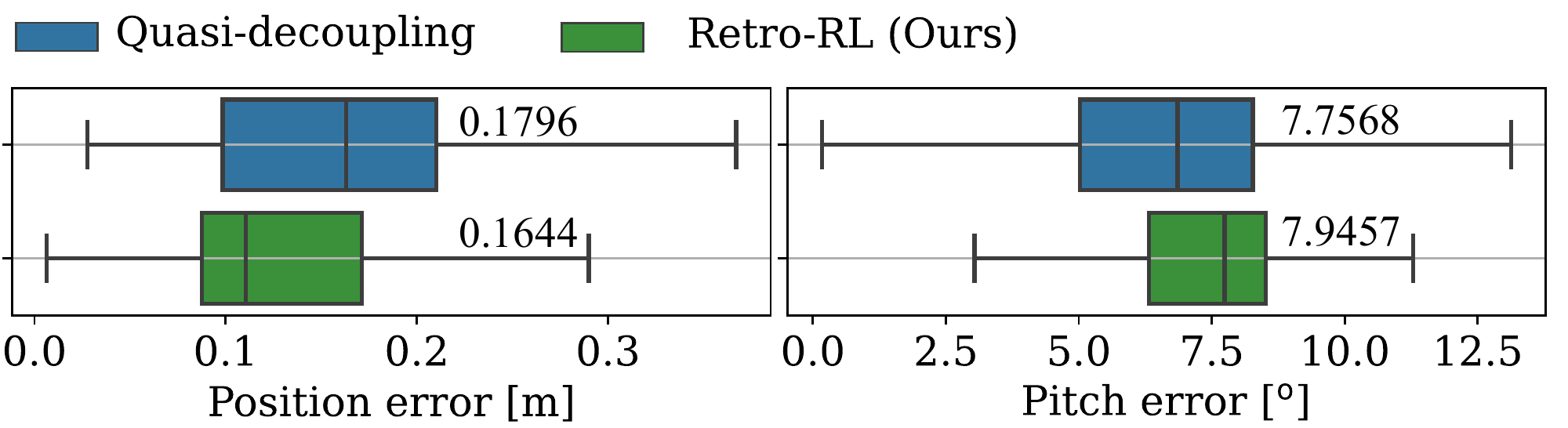}
	\end{subfigure}
	\captionsetup{font=footnotesize}
    \caption{Comparison of pose tracking during wall-climbing operation. The box plot shows position and pitch errors of the CAROS-Q controlled with the quasi-decoupling controller (blue) and Retro-RL (green). The values beside each box plots are RMSEs over time of the corresponding controller.}
	\label{figure:wallclimbing_nominal}
\end{figure}

\subsubsection{\textbf{Wall-Climbing}}
As a final task, the wall-climbing maneuver was performed by commanding the drone to change its pose in mid-air, approaching the wall, and cleaning it (see Fig.~\ref{figure:wall_climbing}). The result of wall-climbing with the PPO-adopted controller is not reported in this paper, since based on the experiments, the PPO-adopted controller fails to get close to the wall because of its instability. Therefore, only the nominal controller is compared with the proposed work for the wall-climbing experiment.

Fig.~\ref{figure:wallclimbing_nominal} compares the pose tracking of CAROS-Q controlled with the nominal quasi-decoupling and Retro-RL during the wall-climbing operation. The result agrees with the pose following result in Fig.~\ref{figure:pose_tracking_comparison}, where Retro-RL significantly decreased the position error. A quantitative analysis of the position error data using a two-sample $t$-Test shows a $p$-value of 0.015, which validates the improvement with a $5\%$ significance level. However, the pitch error was not improved. We hypothesize that it was because the target pitch ($\theta_{des}$) was sub-optimal and Retro-RL figured out better $\theta_{des}$ for more accurate position tracking.

\section{Conclusion and Future Work} \label{section:conclusion}
In this paper, Retro-RL, a novel control framework for combining the advantages of a nominal controller derived from prior physics of a system and a policy learned using a deep RL algorithm, is proposed. Through Lyapunov stability theorem, the Retro-RL framework is guaranteed to be globally, uniformly, and asymptotically stable for unperturbed system and can be ultimately bounded by knowing the perturbation properties. In real-world experiments, the proposed algorithm qualitatively and quantitatively outperformed the stand-alone nominal and learning-based controllers on the CAROS-Q, a tilting-hexarotor drone platform. Overall, this work is expected to open directions for future research in deep RL, particularly for preventing catastrophic failure due to the uncertainties of neural networks. For future work, we plan to extend the proposed method to deal with adverse state estimation with an onboard sensing system and incorporate the rotor dynamics to achieve higher energy efficiency.

\bibliographystyle{IEEEtran}
\bibliography{./root,./IEEEabrv}

\begin{thebibliography}{10}
\providecommand{\url}[1]{#1}
\csname url@rmstyle\endcsname
\providecommand{\newblock}{\relax}
\providecommand{\bibinfo}[2]{#2}
\providecommand\BIBentrySTDinterwordspacing{\spaceskip=0pt\relax}
\providecommand\BIBentryALTinterwordstretchfactor{4}
\providecommand\BIBentryALTinterwordspacing{\spaceskip=\fontdimen2\font plus
\BIBentryALTinterwordstretchfactor\fontdimen3\font minus
  \fontdimen4\font\relax}
\providecommand\BIBforeignlanguage[2]{{%
\expandafter\ifx\csname l@#1\endcsname\relax
\typeout{** WARNING: IEEEtran.bst: No hyphenation pattern has been}%
\typeout{** loaded for the language `#1'. Using the pattern for}%
\typeout{** the default language instead.}%
\else
\language=\csname l@#1\endcsname
\fi
#2}}

\bibitem{youn2021collision}
W.~Youn, H.~Ko, H.~Choi, I.~Choi, J.-H. Baek, and H.~Myung, ``Collision-free
  autonomous navigation of a small {UAV} using low-cost sensors in {GPS}-denied
  environments,'' \emph{International Journal of Control, Automation and
  Systems}, vol.~19, no.~2, pp. 953--968, 2021.

\bibitem{kim2017robust}
S.~Kim, S.~Choi, H.~Kim, J.~Shin, H.~Shim, and H.~J. Kim, ``Robust control of
  an equipment-added multirotor using disturbance observer,'' \emph{IEEE
  Transactions on Control Systems Technology}, vol.~26, no.~4, pp. 1524--1531,
  2017.

\bibitem{myeong2018development}
W.~Myeong and H.~Myung, ``Development of a wall-climbing drone capable of
  vertical soft landing using a tilt-rotor mechanism,'' \emph{IEEE Access},
  vol.~7, pp. 4868--4879, 2018.

\bibitem{myeong2019development}
W.~Myeong, S.~Jung, B.~Yu, T.~Chris, S.~Song, and H.~Myung, ``Development of
  wall-climbing unmanned aerial vehicle system for micro-inspection of
  bridges,'' in \emph{Proc. Workshop on The Future of Aerial Robotics:
  Challenges and Opportunities, International Conference on Robotics and
  Automation (ICRA)}, 2019, pp. 20--24.

\bibitem{lee2021caros}
H.~Lee, B.~Yu, C.~Tirtawardhana, C.~Kim, M.~Jeong, S.~Hu, and H.~Myung,
  ``{CAROS-Q}: Climbing aerial robot system adopting rotor offset with a
  quasi-decoupling controller,'' \emph{IEEE Robotics and Automation Letters},
  vol.~6, no.~4, pp. 8490--8497, 2021.

\bibitem{kamel2018voliro}
M.~Kamel, S.~Verling, O.~Elkhatib, C.~Sprecher, P.~Wulkop, Z.~Taylor,
  R.~Siegwart, and I.~Gilitschenski, ``The {V}oliro omniorientational
  hexacopter: An agile and maneuverable tiltable-rotor aerial vehicle,''
  \emph{IEEE Robotics \& Automation Magazine}, vol.~25, no.~4, pp. 34--44,
  2018.

\bibitem{falanga2018foldable}
D.~Falanga, K.~Kleber, S.~Mintchev, D.~Floreano, and D.~Scaramuzza, ``The
  foldable drone: A morphing quadrotor that can squeeze and fly,'' \emph{IEEE
  Robotics and Automation Letters}, vol.~4, no.~2, pp. 209--216, 2018.

\bibitem{kim2021morphing}
C.~Kim, H.~Lee, M.~Jeong, and H.~Myung, ``A morphing quadrotor that can
  optimize morphology for transportation,'' in \emph{Proc. IEEE/RSJ
  International Conference on Intelligent Robots and Systems (IROS)}, 2021, pp.
  9683--9689.

\bibitem{allenspach2020design}
M.~Allenspach, K.~Bodie, M.~Brunner, L.~Rinsoz, Z.~Taylor, M.~Kamel,
  R.~Siegwart, and J.~Nieto, ``Design and optimal control of a tiltrotor
  micro-aerial vehicle for efficient omnidirectional flight,''
  \emph{International Journal of Robotics Research}, vol.~39, no. 10-11, pp.
  1305--1325, 2020.

\bibitem{davis2018aerodynamic}
E.~B. Davis, \emph{Aerodynamic {F}orce {I}nteractions and {M}easurements for
  {M}icro {Q}uadrotors}.\hskip 1em plus 0.5em minus 0.4em\relax School of
  Information Technology and Electrical Engineering, University of Queensland,
  Australia, 2018.

\bibitem{kocer2018centralized}
B.~B. Kocer, T.~Tjahjowidodo, and G.~G.~L. Seet, ``Centralized predictive
  ceiling interaction control of quadrotor {VTOL UAV},'' \emph{Aerospace
  Science and Technology}, vol.~76, pp. 455--465, 2018.

\bibitem{lee2020aerial}
D.~Lee, H.~Seo, I.~Jang, S.~J. Lee, and H.~J. Kim, ``Aerial manipulator pushing
  a movable structure using a {DOB}-based robust controller,'' \emph{IEEE
  Robotics and Automation Letters}, vol.~6, no.~2, pp. 723--730, 2020.

\bibitem{hwangbo2017control}
J.~Hwangbo, I.~Sa, R.~Siegwart, and M.~Hutter, ``Control of a quadrotor with
  reinforcement learning,'' \emph{IEEE Robotics and Automation Letters},
  vol.~2, no.~4, pp. 2096--2103, 2017.

\bibitem{lambert2019low}
N.~O. Lambert, D.~S. Drew, J.~Yaconelli, S.~Levine, R.~Calandra, and K.~S.
  Pister, ``Low-level control of a quadrotor with deep model-based
  reinforcement learning,'' \emph{IEEE Robotics and Automation Letters},
  vol.~4, no.~4, pp. 4224--4230, 2019.

\bibitem{molchanov2019sim}
A.~Molchanov, T.~Chen, W.~H{\"o}nig, J.~A. Preiss, N.~Ayanian, and G.~S.
  Sukhatme, ``Sim-to-(multi)-real: Transfer of low-level robust control
  policies to multiple quadrotors,'' in \emph{Proc. IEEE/RSJ International
  Conference on Intelligent Robots and Systems (IROS)}, 2019, pp.
  59--66\color{black}.

\bibitem{penicka2022learning}
R.~\color{black}Penicka, Y.~Song, E.~Kaufmann, and D.~Scaramuzza, ``Learning
  minimum-time flight in cluttered environments,'' \emph{arXiv preprint
  arXiv:2203.15052}, 2022\color{black}.

\bibitem{lee2021low}
H.~Lee, M.~Jeong, C.~Kim, H.~Lim, C.~Park, S.~Hwang, and H.~Myung, ``Low-level
  pose control of tilting multirotor for wall perching tasks using
  reinforcement learning,'' in \emph{Proc. IEEE/RSJ International Conference on
  Intelligent Robots and Systems (IROS)}, 2021, pp. 9669--9676.

\bibitem{gharieb2001fuzzy}
W.~Gharieb and G.~Nagib, ``Fuzzy intervention in {PID} controller design,'' in
  \emph{Proc. IEEE International Symposium on Industrial Electronics (Cat. No.
  01TH8570)}, vol.~3, 2001, pp. 1639--1643.

\bibitem{fujii1991neural}
T.~Fujii and T.~Ura, ``Neural-network-based adaptive control systems for
  {AUV}s,'' \emph{Engineering Applications of Artificial Intelligence}, vol.~4,
  no.~4, pp. 309--318, 1991.

\bibitem{bauersfeld2021neurobem}
L.~Bauersfeld, E.~Kaufmann, P.~Foehn, S.~Sun, and D.~Scaramuzza, ``Neuro{BEM}:
  Hybrid aerodynamic quadrotor model,'' in \emph{Proc. Robotics: Science and
  Systems (RSS)}, 2021.

\bibitem{bodie2019omnidirectional}
K.~Bodie, M.~Brunner, M.~Pantic, S.~Walser, P.~Pf{\"a}ndler, U.~Angst,
  R.~Siegwart, and J.~Nieto, ``An omnidirectional aerial manipulation platform
  for contact-based inspection,'' in \emph{Proc. Robotics: Science and Systems
  (RSS)}, 2019.

\bibitem{torrente2021data}
G.~Torrente, E.~Kaufmann, P.~F{\"o}hn, and D.~Scaramuzza, ``Data-driven {MPC}
  for quadrotors,'' \emph{IEEE Robotics and Automation Letters}, vol.~6, no.~2,
  pp. 3769--3776, 2021.

\bibitem{levine2020offline}
S.~Levine, A.~Kumar, G.~Tucker, and J.~Fu, ``Offline reinforcement learning:
  Tutorial, review, and perspectives on open problems,'' \emph{NeurIPS
  Tutorial}, 2020.

\bibitem{song2020learning}
Y.~Song and D.~Scaramuzza, ``Learning high-level policies for model predictive
  control,'' in \emph{Proc. IEEE/RSJ International Conference on Intelligent
  Robots and Systems (IROS)}, 2020, pp. 7629--7636.

\bibitem{song2021policy}
------, ``Policy search for model predictive control with application to agile
  drone flight,'' \emph{IEEE Transactions on Robotics}, 2021.

\bibitem{kim2019improving}
J.~W. Kim, H.~Shim, and I.~Yang, ``On improving the robustness of reinforcement
  learning-based controllers using disturbance observer,'' in \emph{Proc. IEEE
  Conference on Decision and Control (CDC)}, 2019.

\bibitem{guha2020mrac}
A.~Guha and A.~Annaswamy, ``{MRAC-RL}: A framework for on-line policy
  adaptation under parametric model uncertainty,'' in \emph{Proc. Learning for
  Dynamics and Control Conference (L4DC)}, 2021.

\bibitem{kullback1951information}
S.~Kullback and R.~A. Leibler, ``On information and sufficiency,'' \emph{The
  annals of mathematical statistics}, vol.~22, no.~1, pp. 79--86, 1951.

\bibitem{rotors}
\BIBentryALTinterwordspacing
``{F100 T-}motor specifications,'' accessed on 2022.02.18. [Online]. Available:
  \url{https://store.tmotor.com/goods.php?id=1177}
\BIBentrySTDinterwordspacing

\bibitem{robotis}
\BIBentryALTinterwordspacing
``{XM540-W270-T/R} manual,'' accessed on 2022.02.18. [Online]. Available:
  \url{https://emanual.robotis.com/docs/en/dxl/x/xm540-w270/}
\BIBentrySTDinterwordspacing

\bibitem{agarap2018deep}
A.~F. Agarap, ``Deep learning using rectified linear units ({R}e{LU}),''
  \emph{arXiv preprint arXiv:1803.08375}, 2018.

\bibitem{kingma2014adam}
D.~P. Kingma and J.~Ba, ``Adam: A method for stochastic optimization,'' in
  \emph{Proc. International Conference on Learning Representations (ICLR)},
  2015.

\bibitem{tobin2017domain}
J.~Tobin, R.~Fong, A.~Ray, J.~Schneider, W.~Zaremba, and P.~Abbeel, ``Domain
  randomization for transferring deep neural networks from simulation to the
  real world,'' in \emph{Proc. IEEE/RSJ International Conference on Intelligent
  Robots and Systems (IROS)}, 2017, pp. 23--30.

\bibitem{mnih2016asynchronous}
V.~Mnih, A.~P. Badia, M.~Mirza, A.~Graves, T.~Lillicrap, T.~Harley, D.~Silver,
  and K.~Kavukcuoglu, ``Asynchronous methods for deep reinforcement learning,''
  in \emph{Proc. International Conference on Machine Learning (ICML)}, 2016,
  pp. 1928--1937.

\bibitem{schulman2017proximal}
J.~Schulman, F.~Wolski, P.~Dhariwal, A.~Radford, and O.~Klimov, ``Proximal
  policy optimization algorithms,'' \emph{arXiv preprint arXiv:1707.06347},
  2017.

\bibitem{makoviychuk2021isaac}
V.~Makoviychuk, L.~Wawrzyniak, Y.~Guo, M.~Lu, K.~Storey, M.~Macklin,
  D.~Hoeller, N.~Rudin, A.~Allshire, A.~Handa, \emph{et~al.}, ``Isaac {G}ym:
  High performance {GPU}-based physics simulation for robot learning,''
  \emph{NeurIPS Track on Datasets and Benchmarks}, 2021.

\bibitem{rudin2021learning}
N.~Rudin, D.~Hoeller, P.~Reist, and M.~Hutter, ``Learning to walk in minutes
  using massively parallel deep reinforcement learning,'' in \emph{Proc.
  Conference on Robot Learning (CoRL)}, 2021.

\end{thebibliography}

\end{document}